\documentclass[times,twocolumn,final]{article}

\usepackage{amssymb}
\usepackage{latexsym}
\usepackage{graphicx}
\usepackage[style=numeric,backend=biber,bibencoding=utf-8]{biblatex}
\usepackage{float}

\usepackage{url}
\usepackage{xcolor}

\usepackage{hyperref}
\usepackage{subcaption} 
\usepackage[switch,pagewise]{lineno}

\usepackage[cachedir=.,frozencache=true]{minted}
\setminted{fontsize=\small,baselinestretch=1}
\usepackage{mathtools}
\usepackage{array}

\begin{document}

\title{Partial 3D Object Retrieval using Local Binary QUICCI Descriptors and Dissimilarity Tree Indexing}

\author{Bart Iver van Blokland \and Theoharis Theoharis}
\date{July 2021}

\maketitle

\begin{abstract}
A complete pipeline is presented for accurate and efficient partial 3D object retrieval based on Quick Intersection Count Change Image (QUICCI) binary local descriptors and a novel indexing tree. It is shown how a modification to the QUICCI query descriptor makes it ideal for partial retrieval. An indexing structure called Dissimilarity Tree is proposed which can significantly accelerate searching the large space of local descriptors; this is applicable to QUICCI and other binary descriptors. The index exploits the distribution of bits within descriptors for efficient retrieval. The retrieval pipeline is tested on the artificial part of SHREC'16 dataset with near-ideal retrieval results. 
\end{abstract}  

\section{Introduction}

There exist many circumstances in which it is desirable to determine which larger object a smaller surface patch belongs to; occlusions and missing parts can result in this problem. This problem is known as \textit{Partial 3D Object Retrieval}, and a number of methods have been proposed to date which address it \cite{godilRangeScansBased2015} \cite{liuSurveyPartialRetrieval2013} \cite{savelonasOverviewPartial3D2015a} and finds application in areas such as archaeology \cite{sfikasPartialMatching3D2016} \cite{duAutomaticPositioningAlgorithm2016}. 

One successful strategy for partial 3D object retrieval is using the descriptiveness of local shape descriptors, as local surface similarity tends to be maintained when other parts of the object are missing. A problem with retrieval using local shape descriptors is the large number of such descriptors that are generated, potentially one for every vertex. This can be somewhat counteracted by using a salient point detector, but then the retrieval quality is affected by the consistency of this detector. An efficient indexing scheme is therefore called for.

To address this issue, a complete pipeline is presented in this paper which is capable of indexing and retrieving arbitrary 3D objects based on partial queries. Under ideal circumstances the system can achieve near perfect retrieval, even with low degrees of partiality, within reasonable time constraints. 

The pipeline utilises the recently introduced Quick Intersection Count Change Image (QUICCI) descriptor \cite{blokland2020an} whose binary nature makes it storage-efficient and fast to compare.

As part of this complete partial retrieval pipeline, the following novelties are introduced:
\begin{enumerate}
    \item An indexing scheme called ``Dissimilarity Tree'' for efficiently retrieving binary descriptors, especially nearest neighbours with high Hamming distance.
    \item An algorithm for accelerating partial 3D object retrieval using the aforementioned indexing scheme.
    \item An adaptation of the QUICCI descriptor generation process to greatly improve its performance in partial 3D object retrieval applications.
\end{enumerate}

The primary descriptor and distance function used in this pipeline, along with relevant background, is given in Section \ref{sec:related_work}. The pipeline is described at a high level in Section \ref{sec:retrieval_pipeline}, and the two other main contributions which are used in this pipeline are detailed in Sections \ref{sec:dissimilarity_tree} and \ref{sec:quicci_modification}. The various methods are evaluated in Section \ref{sec:evaluation}, and some aspects of those are discussed in Section \ref{sec:Discussion}.

\section{Related Work}
\label{sec:related_work}

The problem of Partial Object Retrieval has to date received significant attention, both using global and using local descriptors. A number of binary descriptor indexing strategies have also been proposed. This work builds upon the QUICCI local descriptor and the Weighted Hamming distance function, which are discussed in detail.

\subsection{Partial Object Retrieval}

Partial Object Retrieval approaches presented to date can  be divided into three main categories; Bag of Visual Words (BoVW) based, View-based, and Part-based  \cite{liuSurveyPartialRetrieval2013} \cite{savelonasOverviewPartial3D2015a}. Other methods also exist, addressing particular applications such as CAD shape retrieval \cite{baiDesignReuseOriented2010}.

BoVW based methods use local feature descriptors to exploit that from the perspective of a local neighbourhood, shapes in a query remain similar to those in the corresponding object in a database. Lavoué et al. \cite{lavoueCombinationBagofwordsDescriptors2012} segment a surface into small patches, and compute a codebook for each patch. Object classification is subsequently done by matching new patches against words in the codebook. Savelonas et al. \cite{savelonasFisherEncodingAdaptive2016} propose an extension to the FPFH \cite{rusu2009fast} descriptor called \textit{dFPFH}, which is used for both local and global matching in their retrieval pipeline. Ohbuchi et al. \cite{ryutarouohbuchiSalientLocalVisual2008a} combine the BoVW and the view-based paradigm by computing a bag of features over range images of an object rendered from different viewpoints, and comparing features of a query against those in a codebook. More recently, Dimou et al. \cite{dimouFast3DScene2020} used features computed from patches from segmented depth images.

Part-based methods use segmentation to divide a shape into smaller distinct patches, computing one feature vector for each of them, then match these against a database of feature vectors from other parts. Agathos et al. \cite{agathos3DArticulatedObject2010} used a graph of segmented parts to locate objects with a similar structure. Tierny et al. \cite{tiernyPartial3DShape2009} used Reeb graphs for both segmenting and encoding relationships between surface patches for partial object retrieval. Furuya et al. \cite{furuyaRandomizedSubVolumePartitioning2015} proposed the RSVP algorithm, which partitions an object into random cuboid volumes, and describes each partition as a binary string, against which other parts can be matched. They, and others, \cite{furuyaLearningPartinwholeRelation2018} later utilised a Siamese-like network pair to project handcrafted features extracted from segmented parts into a common feature space, allowing for fast surface patch comparison.

View-based methods are able to adapt work on image matching and recognition to 3D shapes. Examples of such methods include work by Axenopoulos et al. \cite{dutagaciSHREC2009Shape2009}, who proposed a combination of several features computed from object silhouette outlines to create the Compact Multi-View Descriptor (CMVD). The SIFT local feature \cite{lowe2004distinctive} is used by several methods on images extracted from 3D shapes  \cite{dutagaciSHREC2009Shape2009} \cite{ohbuchiScaleweightedDenseBag2009}. One specific example is work by Sfikas et al. \cite{sfikasPartialMatching3D2016}, where the Authors used the PANORAMA descriptor \cite{papadakisPANORAMA3DShape2010} for matching parts of archaeological to complete objects. More recently, Tashiro et al. \cite{godilRangeScansBased2015} proposed a pipeline relying extensively on the SURF  \cite{bay2008speeded} local feature descriptor. 

In the SHREC'16 Partial 3D Object Retrieval track \cite{Pratikakis_2016}, a number of additional view-based methods were introduced. Aono et al. presented three variant methods which each encoded KAZE features \cite{alcantarilla2012kaze} extracted from different object views with Vector of Locally Aggregated Descriptors (VLAD) \cite{jegou2010aggregating}, Gaussian of Local Distribution (GOLD) \cite{SERRA201522}, and Fisher Vectors (FV) \cite{perronnin2010improving}. Pickup et al. used a variant of the view-based method by Lian et al. \cite{lian2013cm}, using rendered views and SIFT descriptors to find matching points.

\subsection{The QUICCI Local Binary Descriptor}

\begin{figure}
    \centering
    \includegraphics[width=8.0cm]{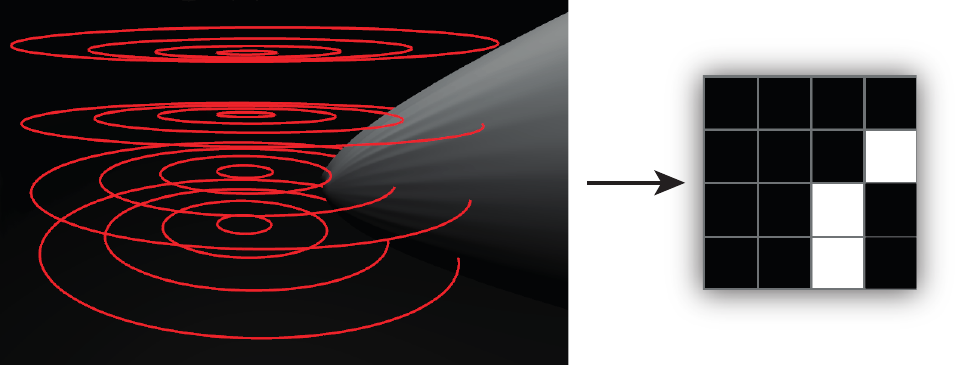}
    \caption{\small Visualisation of a 4x4-bit QUICCI descriptor construction along with the corresponding generated descriptor. White pixels in the descriptor image correspond to a bit value set to 1 (i.e. intersection counts changed), and 0 otherwise.}
    \label{fig:quicci_construction}
\end{figure}

The Quick Intersection Count Change Image, proposed by van Blokland et al. \cite{blokland2020an}, is a binary descriptor which captures changes in intersection counts between circles and an object's surface. These circles are laid out in layers, where each layer contains circles with linearly increasing radii. A visualisation of this structure can be seen in Figure \ref{fig:quicci_construction}.

As can be seen in the Figure, a grid of 4x4 circles is intersecting a 3D surface. A total of 5 circles intersect with this surface, and the remainder do not. To its right the corresponding QUICCI descriptor is shown, where black pixels indicate a bit value of 0, and white a value of 1. Note that each bit has a corresponding circle, where the bit in the bottom left corner of the descriptor is mapped to the innermost circle on the bottom layer.

Each bit in the descriptor denotes whether the number of intersections between the circle corresponding to that bit, and the circle one step smaller on the same layer, has changed. In the Figure, the bottom right 2x2 bits all have corresponding circles which intersect the object surface, which causes a response in the bottom half of column 3, but not in the bottom half of column 4, as the intersection counts did not change.

The resulting descriptor will commonly show outlines of surfaces present near the oriented point around which the descriptor is generated. This point lies at the centre of the grid of circles, which on the descriptor corresponds to the grid point closest to the arrow's head in Figure \ref{fig:quicci_construction}. 

\subsection{Weighted Hamming Distance Function}

There exist two possible bit errors when comparing a pair of binary descriptors (corresponding to a query shape and a target shape from a database, respectively) using a bitwise distance function such as Hamming distance. A type A error occurs when a bit set to 1 in the query is set to 0 in the target, and a type B error represents the case where a bit set to 0 in the query is set to 1 in the target. The Hamming distance function considers both of these bit errors as equivalent in importance.

Meanwhile, the Weighted Hamming distance function proposed by van Blokland et al. \cite{blokland2020an} observes that it may not always be desirable to weigh both types of bit errors equally. In the case of QUICCI descriptors, bits set to 1 represent surface outlines. A good match must also contain these bits, but may also include others due to responses from other geometry. For QUICCI descriptors the type A error is therefore more important than the type B error. 

The Weighted Hamming distance function normalises the contribution of each bit error type by the total number of such errors that \textit{can} occur, thereby weighting the importance of each bit error type equally as a group. Thus the Weighted Hamming distance function is asymmetric. In a sparse descriptor, this implies that a type A error is weighted much more than a type B error. The distance function is listed in Equation \ref{eq:weightedHamming}.

\begin{align}
\begin{split}
    \label{eq:weightedHamming}
    \delta_{WH}(D_q, D_t) &=  \frac {\sum_{r = 1}^{N} \sum_{c = 1}^{N} (D_{q}[r,c](1 - D_t[r,c]))} {max(\sum_{r = 1}^{N} \sum_{c = 1}^{N} D_{q}[r,c], 1)} \\
    &+ \frac {\sum_{r = 1}^{N} \sum_{c = 1}^{N} ((1 - D_{q}[r,c])D_t[r,c])} {max(N - \sum_{r = 1}^{N} \sum_{c = 1}^{N} D_{q}[r,c], 1)}
\end{split}
\end{align}

Where $D_q$ and $D_t$ are respectively the query and target descriptors being compared, $D[r,c]$ represents the bit at row $r$ and column $c$ of descriptor $D$, and the size of the descriptor is $N x N$ bits. 

Experiments by van Blokland et al. showed that using the Weighted Hamming distance function resulted in improved retrieval performance relative to Hamming distance of QUICCI descriptors when additive noise was applied.

\label{subsec:binary_indexes_background}
\subsection{Indexing of Binary Descriptors}

The need for indexing binary descriptors commonly arises in algorithms representing shape features as binary descriptors, but also in other fields such as dimensionality reduction through Locality-Sensitive Hashing (LSH) \cite{Har-Peled_Indyk_Motwani_2012}. Some popular methods utilising LSH include Minhash proposed by Broder et al. \cite{broder1997on} \cite{broder2000identifying} and Simhash \cite{sadowski2007simhash} by Sadowski et al.

There exist a number of methods which produce and compare binary shape features, such as BRIEF by Calonder et al. \cite{calonder2010brief}, an extension called ORB by Rublee et al. \cite{rublee2011orb}, and BRISK by Leutenegger et al. \cite{leutenegger2011brisk}. A binary descriptor which has specifically been proposed for 3D point matching is B-SHOT by Prakhya et al. \cite{prakhya2015b}, and the aforementioned QUICCI descriptor by van Blokland et al. \cite{blokland2020an}.

Local binary descriptors are often produced in large quantities, which raises the need for acceleration structures capable of efficiently locating nearest neighbours in Hamming space. A number of methods have been proposed for this purpose \cite{brodal1996approximate} \cite{brin1995near} \cite{arslan2004dictionary} \cite{maai2007text}. Unfortunately, these initial attempts only support short descriptors, are limited to the retrieval of neighbours up to a Hamming distance of 2, or both. This significantly limits their applicability.

More recent work includes the Multi-Index Hashing (MIH) algorithm proposed by Norouzi et al. \cite{norouzi2012fast}, which subdivides descriptors into regions, building inverted hash tables for each subdivision. Chappell et al.  \cite{chappell2015approximate} proposed a similar approach, instead using inverted lists. An improved variation of MIH was presented by Reina et al. \cite{reina2017an}, utilising a prefix tree to store the index itself, and a separate hash table for pruning irrelevant branches while querying.

The Hamming Tree proposed by van Blokland et al. \cite{blokland2020an} exploits the notion that descriptors with a low Hamming distance must by necessity have a similar number of bits set to 1. The tree first categorises descriptors by the total number of 1 bits, then divides descriptors into regions, categorising them by the number of bits set to 1 within each region.

Unfortunately, the previously introduced binary indexes typically assume that the nearest neighbour to a query in the database has a low Hamming distance, which is not a property which can be assumed consistently. This issue is particularly significant for the application of QUICCI descriptors on the problem of partial retrieval. The discussed indexing strategies tend to scale poorly with increasing distance from a descriptor to its nearest neighbour, which makes their application intractable when this distance is high. 

Moreover, they cannot be adapted to use the Weighted Hamming distance function, which is a highly desirable property for the application of QUICCI descriptors.

\section{Partial Retrieval Pipeline}
\label{sec:retrieval_pipeline}

\begin{figure*}
    \centering
    \begin{subfigure}[t]{16cm}
        \includegraphics[width=16cm]{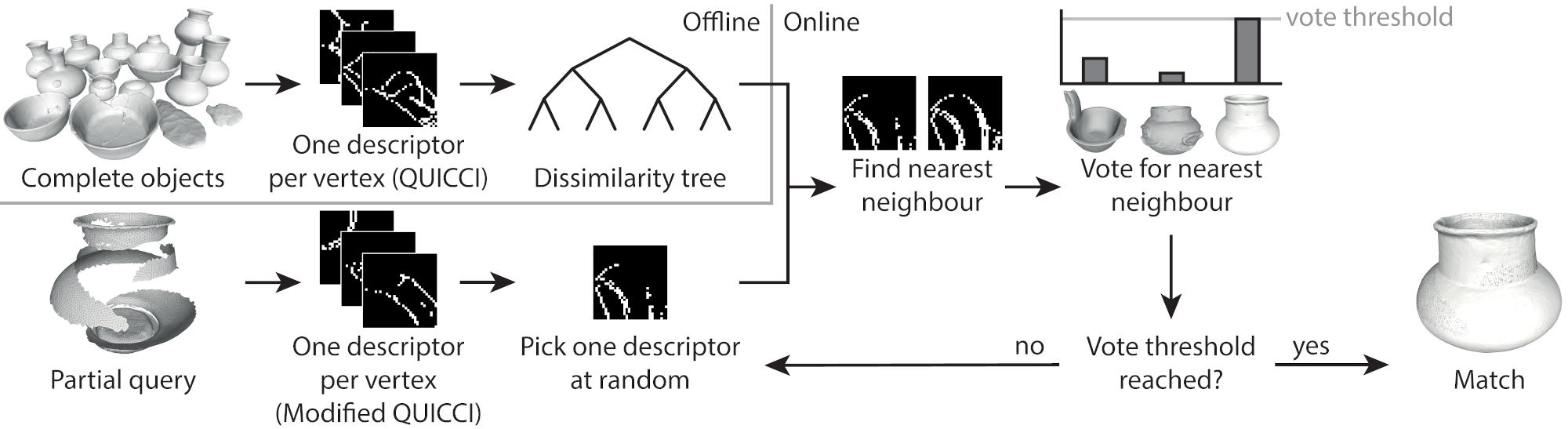}
    \end{subfigure}
    \caption{\small An overview of the proposed partial retrieval pipeline. A dissimilarity tree is constructed offline over the QUICCI descriptors of a set of complete objects. Querying these objects with a partial object involves computing \textit{modified} QUICCI descriptors for each vertex, iteratively, selecting one such query descriptor at random, determining the closest indexed descriptor to the one randomly selected, and finally voting for the object from which the nearest neighbour originated. When an object has reached a set number of votes, it is deemed the closest match to the partial query.}
    \label{fig:pipeline_overview}
\end{figure*}

The proposed partial retrieval pipeline consists of an online and offline component, and is visualised in Figure \ref{fig:pipeline_overview}. In the offline phase, one QUICCI descriptor is extracted from each vertex in a set of complete objects. Using these descriptors, a dissimilarity tree is constructed, which is described in detail in Section \ref{sec:dissimilarity_tree}. 

The online phase takes a partial query object as input. In similar fashion to the offline phase, one descriptor is computed for each of the object's vertices, albeit a  \textit{modified} version of the QUICCI descriptor is used here, see Section \ref{sec:quicci_modification}. The voting steps outlined below are then repeated until an object match is found or all vertices have been exhausted.

One descriptor is first selected from the set of query descriptors at random. Using the dissimilarity tree structure, the nearest neighbour in terms of Weighted Hamming Distance is found in the set of descriptors extracted from the complete objects. Next, a vote is cast for the object containing the found nearest neighbour. This process is repeated until an object has received a predefined number of votes (threshold), upon which this object is considered the best match for the partial query. Our evaluation shows that a threshold of 10 votes is sufficient. Additional search results can be obtained by evaluating additional queries until the desired number of other objects reaches the vote threshold.

\begin{figure}
    \centering
    \includegraphics[width=8.0cm]{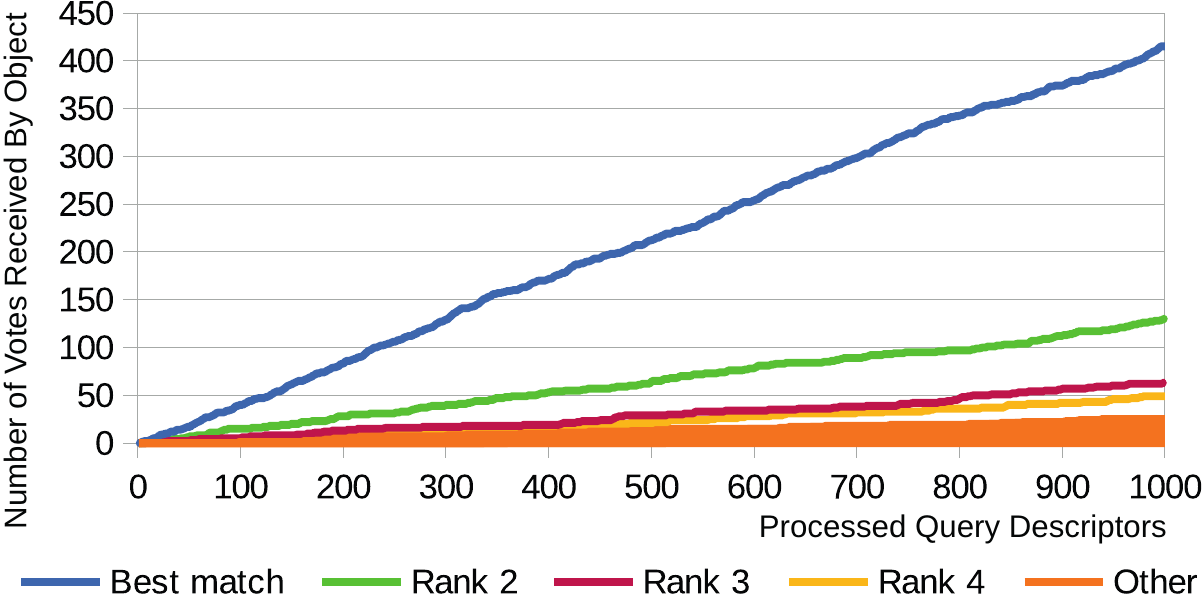}
    \caption{\small A visualisation of vote counts received by objects as more query descriptors are processed, while using the proposed pipeline to locate the nearest neighbour of a single partial query object. Each line represents a single object from the SHREC'16 partial retrieval dataset \cite{Pratikakis_2016}. For example, the best match (blue line) received 300 of the first 700 votes cast. The linear nature of these curves suggests that the probability that a particular object will receive a vote is approximately constant across the voting process.}
    \label{fig:linearOccurrenceProgression}
\end{figure}

The motivation for using a voting threshold to exit the counting process can be seen in Figure \ref{fig:linearOccurrenceProgression}, which shows the number of votes received by different objects with respect to the number of processed query descriptors using the proposed method. As this relationship is approximately linear, it is possible to terminate the search early by using a vote threshold.

\begin{figure}
    \centering
    \includegraphics[width=8.0cm]{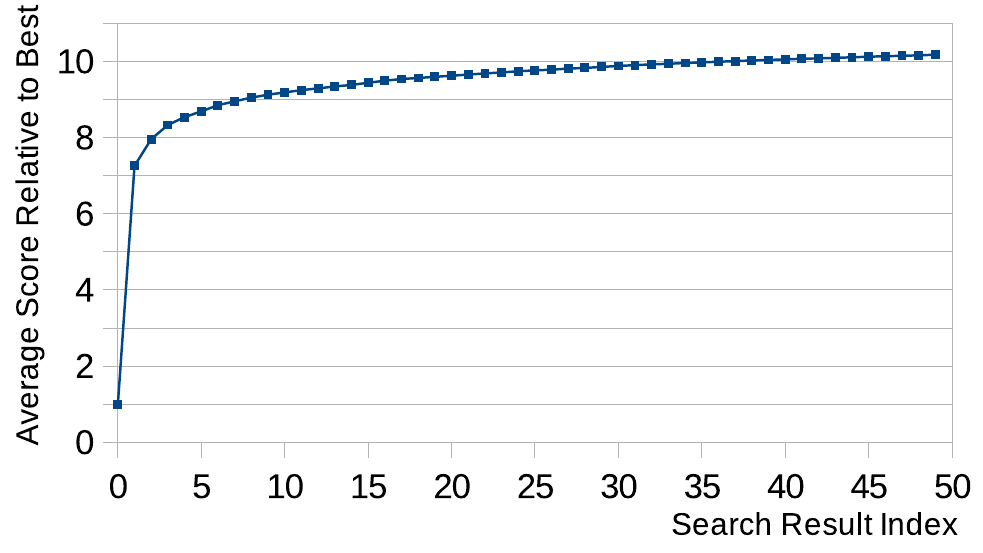}
    \caption{\small Average Weighted Hamming distance of the top 50 descriptor search results (where the top search result has index 0) of 1,000 descriptor queries normalised to the distance score of the top search result. As the average distance between the top and second ranked search result is high, it is unlikely that any result but the best result is relevant to the query.}
    \label{fig:average_score_chart}
\end{figure}

For each randomly selected query descriptor, only the nearest neighbour descriptor is considered. The motivation for not considering other neighbours in addition to the nearest is shown in Figure \ref{fig:average_score_chart}. The Figure shows the average Weighted Hamming distance scores of the closest 50 matches to 1,000 descriptor queries. For legibility, distance scores for each neighbour have been normalised relative to the Weighted Hamming distance of the nearest neighbour (search result index 0) of the same query. The Figure shows that the average distance score to the query descriptor increases between the nearest and second nearest neighbours by over a factor of 7, where further neighbours exhibit similar scores. We therefore conclude only the nearest neighbour to the query is relevant to the retrieval process.

\section{Dissimilarity Tree for Indexing Binary Descriptors}
\label{sec:dissimilarity_tree}

The Hamming Tree \cite{blokland2020an} and other binary descriptor indexing structures proposed in previous work (see Section \ref{subsec:binary_indexes_background}) have the ability to efficiently locate descriptors similar to a given query when the Hamming distance to those matches is generally low. However, in applications where this distance is large, the search time of these methods tends to increase significantly. 

This problem has a high likelihood of occurring when using QUICCI descriptors for partial object retrieval. In this case, a partial query descriptor will generally contain a subset of the bits set compared to those for the correctly matching descriptor of the complete object.

A new tree indexing structure, the \textit{Dissimilarity Tree}, is proposed here which is capable of efficiently retrieving nearest neighbours when the distance to these neighbours is high. The Dissimilarity Tree is also capable of supporting the Weighted Hamming distance function for querying, which has been shown to be superior for QUICCI descriptor ranking \cite{blokland2020an}. The Dissimilarity Tree can be used for arbitrary binary descriptors, but in the remainder of this paper will be explained in the context of QUICCI descriptors. 

The Dissimilarity Tree is a binary tree that exploits the assumption that bits set to 1 in a binary descriptor are not distributed randomly, and aims to cluster descriptors which have similar bits set. It does so by attempting to create subtrees where patterns of bits are consistently set in all contained descriptors. 

When it is known that a specific bit will have a consistent value (0 or 1) across all descriptors in a subtree, it is possible to compute the (Weighted) Hamming distance that will be incurred for that bit for a given query descriptor. The more effectively this can be done, the greater the ability of the search algorithm to prune irrelevant branches.

For each node in the Dissimilarity Tree, two characteristic binary images are computed, one representing the bitwise sum (OR) of all descriptors contained in both subtrees and one representing the bitwise product (AND). These allow a minimum distance to be computed from the query descriptor to all descriptors contained in that particular node, as the sum and product descriptors denote which bits in a subtree are consistently set to 0 and 1, respectively. 

\begin{figure*}
    \centering
    \begin{subfigure}[t]{16cm}
        \includegraphics[width=16cm]{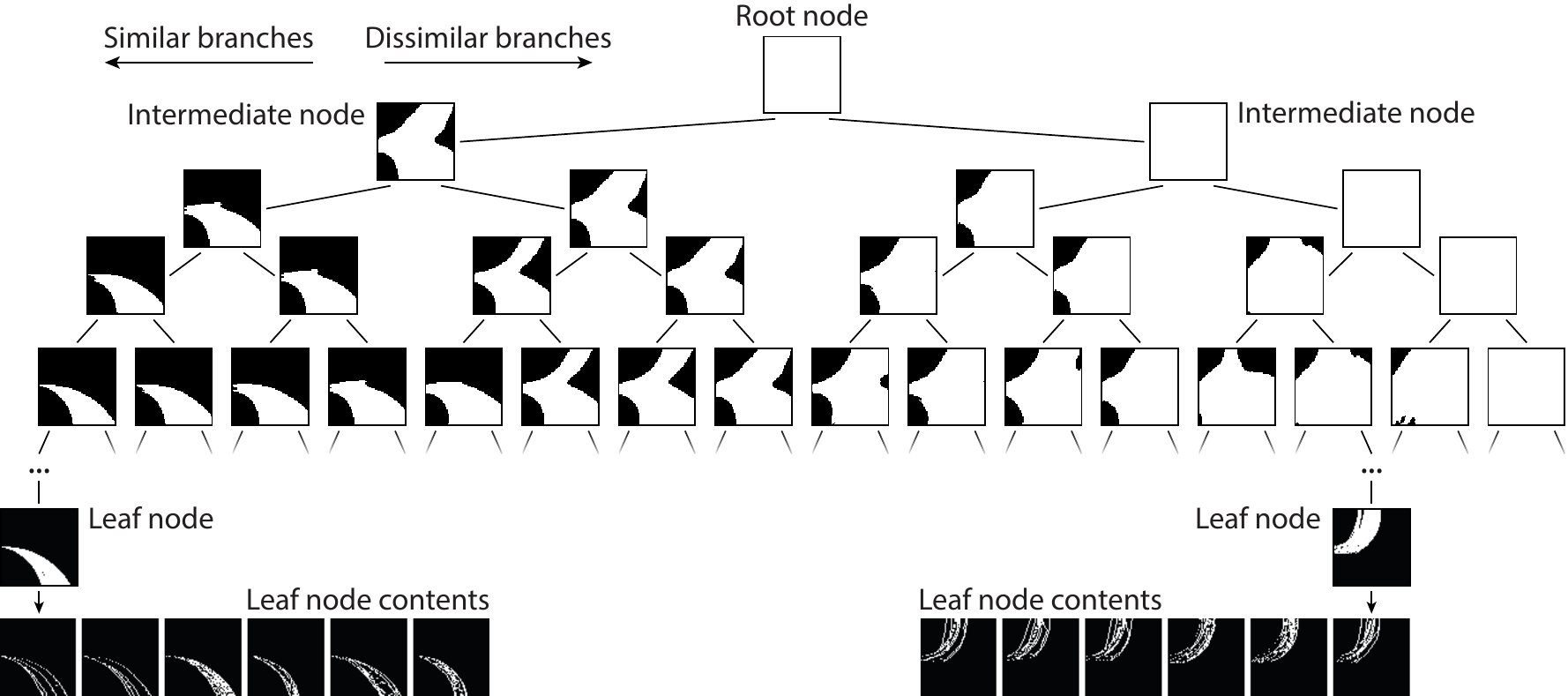}
    \end{subfigure}
    \caption{\small Visualisation showing the four top layers of the Dissimilarity tree generated from all descriptors in the SHREC'16 dataset. Each node is represented by its sum image. All outgoing branches from nodes directed to the left represent the similar branch of that node, and those directed to the right are those that are dissimilar. Two examples of leaf nodes are also shown, along with a subset of the descriptors contained within each.}
    \label{fig:index_visualisation}
\end{figure*}

At each node, the set of descriptors is partitioned into two roughly equal subsets. The \emph{similar} subset contains descriptors which maximise the number of bits consistently set to 0 or 1. The remaining descriptors form the \emph{dissimilar} subset. A visualisation of an example dissimilarity tree can be seen in Figure \ref{fig:index_visualisation}. 

For descriptors which are either highly sparse or dense, the product and sum images respectively do not provide meaningful value to the querying process when the Weighted Hamming distance function is used. Since QUICCI descriptors are highly sparse, we omit the product image. Section \ref{subsec:tree_construction} details the algorithm for constructing a Dissimilarity Tree while Section \ref{subsec:tree_querying} describes the querying process.


\subsection{Dissimilarity Tree Construction}
\label{subsec:tree_construction}

The tree construction algorithm described in this section details how a dissimilarity tree can be constructed from a fixed set of descriptors. However, it should be possible to construct a tree incrementally by recomputing only the affected parts of the tree, in a similar fashion to the heapify algorithm \cite{cormen2009introduction}.

The root of the tree represents the set of all descriptors in the index. The tree construction algorithm divides this set into two approximately equally sized disjoint subsets. The \emph{similar} subset of these contains descriptors where regions of bits set consistently to 0 or 1 are created, by moving descriptors where those bits are set otherwise to the \emph{dissimilar} set. For each of these subsets a new node is created, and the procedure is repeated recursively until the set of descriptors represented by a node is smaller than a threshold. Pseudocode for the construction algorithm is shown in Listing \ref{listing:buildIndexPseudocode}.

\begin{listing}
\begin{minted}[samepage]{python}
def buildDissimilarityTree(node, descriptors):
  if descriptors.length <= maxDescriptorsPerLeaf:
    markAsLeafNode(node)
    return
  similarDescriptors = descriptors.copy()
  dissimilarDescriptors = create_set() # empty
  # Count popularity of set bits in descriptors
  counts = computeBitResponseCounts(descriptors)
  bitOrder = sortAscending(counts)
  
  # Split descriptors into similar and 
  # dissimilar subsets
  for bitIndex in bitOrder:
    for descriptor in similarDescriptors:
      if descriptor[bitIndex] == 1:
        similarDescriptors.remove(descriptor)
        dissimilarDescriptors.insert(descriptor)
  
  # Recurse for similar and dissimilar branches
  node.similarBranch = create_node()
  buildDissimilarityTree(node.similarBranch, 
                         similarDescriptors)
  node.dissimilarBranch = create_node()
  buildDissimilarityTree(node.dissimilarBranch, 
                         dissimilarDescriptors)
  node.productImage = computeAND(descriptors)
  node.sumImage = computeOR(descriptors)
\end{minted}
\caption{\small Pseudocode of the Dissimilarity Tree construction algorithm.}
\label{listing:buildIndexPseudocode}
\end{listing}

\begin{figure}
    \centering
    \includegraphics[width=8.0cm]{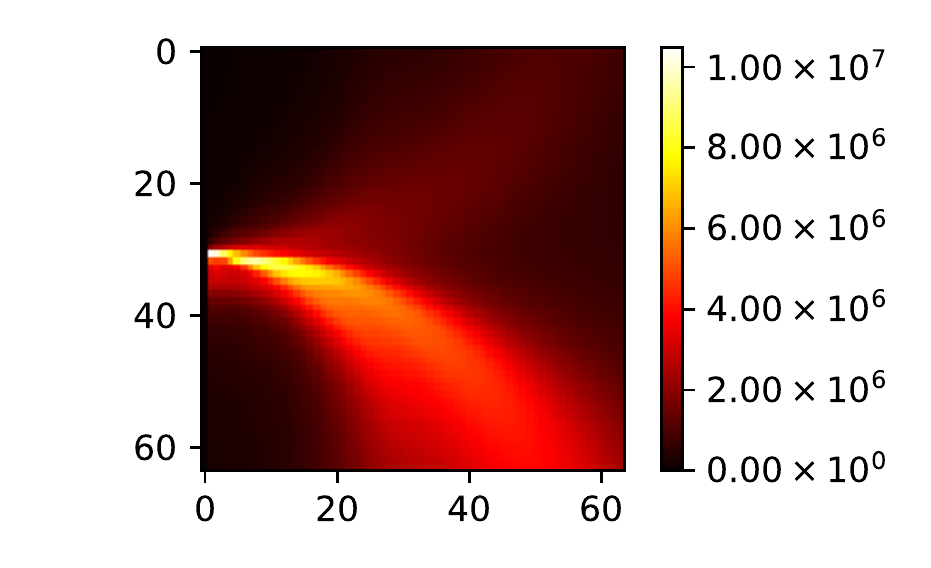}
    \caption{\small Heatmap showing occurrence counts of bits of all descriptors of complete objects in the SHREC'16 partial retrieval dataset.}
    \label{fig:occurrenceHeatmap}
\end{figure}

In order to maximise the size of the region of bits set to a consistent value, the number of descriptors in a given set is determined for which a specific bit is set to the value 1, which will be referred to as the \emph{popularity} of that bit. 
The bit popularity can be visualised in a heatmap, an example of which, computed over the entirety of the SHREC'16 Partial Object Retrieval dataset, is shown in Figure \ref{fig:occurrenceHeatmap}. As can be seen in the Figure, there are areas where bits are frequently set to 1 (middle left), and others less so (top and bottom left, middle right).

The popularity heatmap in Figure \ref{fig:occurrenceHeatmap} represents the occurrence counts for the root node of Figure \ref{fig:index_visualisation}. In the latter Figure, the sum image of the node along the similar branch shows that the less common areas of the heatmap have been cut away as part of the first subdivision, leaving zero-valued areas in the sum image. 

The division strategy starts by computing the aforementioned bit popularity heatmap. Next, bit positions within the descriptor are sorted in order of ascending popularity, the \textit{bitOrder}. 

Next all descriptors of the current node are placed in a the \textit{similarDescriptors} set and the \textit{dissimilarDescriptors} set is initalised to the empty set. Then, for each bit position in \textit{bitOrder} starting from the least popular one, all descriptors are found in \textit{similarDescriptors} which have that particular bit set to 1. These descriptors are moved over to \textit{dissimilarDescriptors}. Moving all descriptors that have a specific bit set may result in a tree which is not perfectly balanced, however, only moving a part of the descriptors which have a particular bit set to 1 does not yield the advantage of that particular bit being set to 0 in the similar node's sum image.

If the node being visited contains a set of descriptors which has fewer descriptors than a set threshold, the subdivision can stop. By constructing a number of indices, it was determined that the value of 32 for this threshold yields optimal execution times for QUICCI descriptors.

\subsection{Dissimilarity Tree Querying}
\label{subsec:tree_querying}

As mentioned previously, the sum and product images of each node allow a minimum distance to be computed between the query descriptor and all descriptors contained within both branches of a particular node. This is possible because the sum and product images by definition represent all descriptors contained within the node having a particular bit set to 0 or 1, respectively. 

If for instance a particular bit in the query descriptor is set to 1, and the sum image of a node has that same bit set to 0, every single descriptor contained within the node will incur a distance penalty at that bit.

By summing up all such universal distances using a distance function, such as Hamming or Weighted Hamming, a minimum distance can be computed between the query descriptor and all descriptors contained in the node being considered, thus allowing to make decisions on culling a subtree. 

The querying algorithm of the dissimilarity tree works in an iterative fashion, and is outlined in Listing \ref{listing:queryIndexPseudocode}. A priority queue is kept of open nodes, sorted by their minimum distance. The queue is initialised to the root node. During each iteration, the node with the lowest minimum distance is removed from the queue. If the node is a leaf node, the list of descriptors contained is searched for matches. If the node is not a leaf node, the minimum distance to the similar and dissimilar branch nodes is computed, and both are inserted into the queue.

Meanwhile, a list of fixed size is kept with the closest matching descriptors found up to that point. The list is sorted by each result's Weighted Hamming distance to the query. When the list of search results has the desired size and the worst result in the list is lower than the minimum distance to the next unvisited open node, all search results have been found and the search can terminate.

\begin{listing}
\begin{minted}[samepage]{python}
def queryDissimilarityTree(query, rootNode, 
                           resultCount):
  openQueue = create_priority_queue()
  openQueue.insert(rootNode)
  searchResults = create_list()
  # While we need more results and nodes
  # that can improve the results exist
  while len(searchResults) < resultCount and
        searchResults[-1].distance > 
        openQueue[0].minDistance):
    node = openQueue.remove(0)
    if isLeafNode(node):
      searchResults.append(node.descriptors)
      searchResults.computeDistanceTo(query)
      searchResults.sortByDistance()
      searchResults.shrinkTo(resultCount)
    else: # node is intermediate node
      node.similarBranch.minDistance
       = minWHDist(query, node.similarBranch)
      node.dissimilarBranch.minDistance
       = minWHDist(query, node.dissimilarBranch)
      # Only consider child nodes if
      # they can improve search results
      if node.similarBranch.minDistance < 
         searchResults[-1].distance:
        openQueue.insert(node.similarBranch)
      if node.dissimilarBranch.minDistance < 
         searchResults[-1].distance:
        openQueue.insert(node.dissimilarBranch)
      
# Compute minimum weighted Hamming distance
def minWHDist(query, node):
  return sum(popcnt(query and not node.sumImage))
         * (len(query) / popcnt(query))
   + sum(popcnt(not query and node.productImage))
     * (len(query) / 
       (len(query) - popcnt(query))
\end{minted}
\caption{\small Pseudocode showing the main steps of the Dissimilarity Tree querying algorithm. The procedure takes a query descriptor, the root of a Dissimilarity Tree, and the desired number of closest search results to retrieve.}
\label{listing:queryIndexPseudocode}
\end{listing}


\section{Adapting QUICCI Descriptors for Partial Retrieval}
\label{sec:quicci_modification}

Partial objects that constitute queries are generally not closed surfaces. They commonly contain surface discontinuities, which we shall call \textit{boundaries}. Thus if the QUICCI descriptor is used on vertices of such a partial object, one would get responses to boundaries, that have no match in the corresponding complete object.

\begin{figure}
    \centering
    \includegraphics[width=8.0cm]{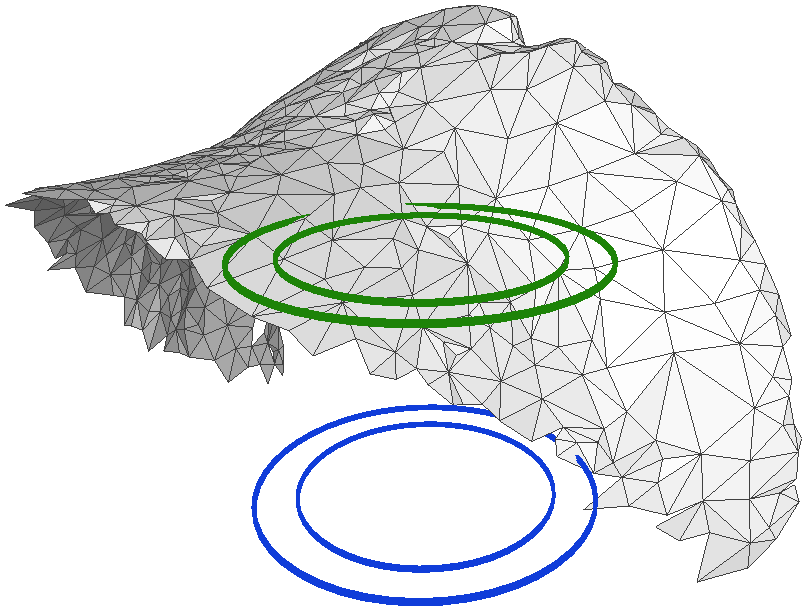}
    \caption{\small A query mesh along with two pairs of hypothetical circles used during the construction of the QUICCI descriptor. The green circles on top indicate an intersection count change of 2, while the blue circle pair on the bottom indicates an intersection count change of 1.}
    \label{fig:singleintersectionfilter}
\end{figure}

An example of a partial query object can be seen in Figure \ref{fig:singleintersectionfilter}, where the intersections count changes of successive circle pairs result in QUICCI descriptor responses. The green ones will result in a response which has a match in the corresponding complete object but the blue ones (across a boundary) will not. Ideally, these boundary responses should be filtered out. 

Fortunately, a slight modification to the computation process of QUICCI descriptors for partial query objects, effectively filters out most interference caused by boundaries. This is based on the fact that boundaries result in intersection count changes by 1, whereas closed surfaces result in intersection count changes by 2. Thus, during QUICCI construction of partial query objects we only record intersection deltas of at least 2.

\begin{figure}
    \centering
    \includegraphics[width=8.0cm]{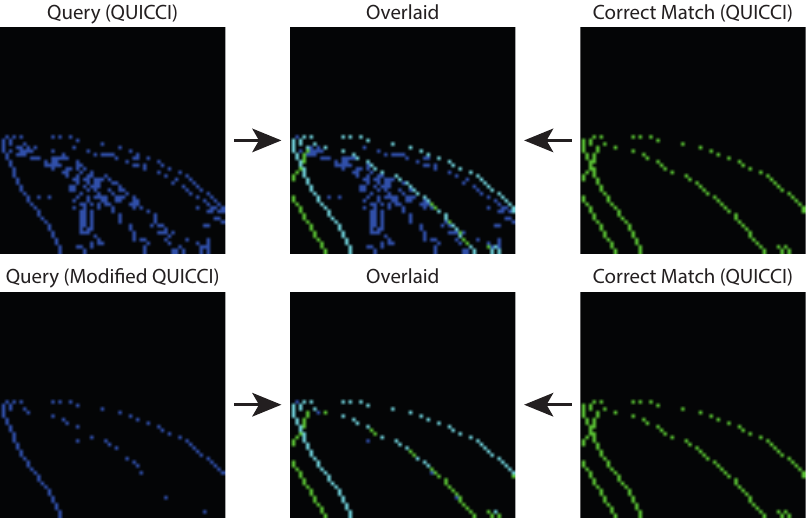}
    \caption{\small QUICCI descriptors for a vertex from a partial query object (left side), and its corresponding matching vertex in the complete object (right side). The top left is an original QUICCI descriptor while the bottom left is a modified QUICCI. Both query-match pairs are shown overlaid on top of each other in the middle, and show a significant reduction in noise in the query descriptor when using the modified QUICCI.}
    \label{fig:modifiedQuicciComparison}
\end{figure}

Figure \ref{fig:modifiedQuicciComparison} shows a comparison between the existing QUICCI descriptor and the proposed modification for partial query objects, along with the matching descriptor for the corresponding complete object. The proposed change filters out nearly all responses induced by boundaries, leaving responses belonging to the shape being queried. Ideally, the response set of the modified query image is a subset of the corresponding set for the complete object.

\section{Evaluation}
\label{sec:evaluation}

The proposed partial retrieval pipeline was evaluated on a set of real 3D object scans. The primary dataset chosen for this purpose is the SHREC'16 Partial Object Retrieval track dataset \cite{Pratikakis_2016}, which consists of a variety of historic artefacts, primarily ceramic pottery. We create additional partial query objects from this dataset to form a new augmented dataset of partial query objects, detailed in Section \ref{subsec:augmentation}. 

The three primary contributions are subsequently evaluated individually. The Dissimilarity Tree is evaluated in Section \ref{subsec:dissimilarity_tree_results}. The proposed QUICCI modification for partial query objects is evaluated for its ability to filter responses to query boundaries in Section \ref{subsec:results_modified_equicci_filtering}, and for its matching capabilities in Section \ref{subsec:results_modified_quicci}. The complete partial retrieval pipeline is finally externally evaluated using the augmented SHREC'16 dataset in Section \ref{subsec:results_pipeline} and on part of the original SHREC'16 query objects in Section \ref{subsec:results_shrec2016}.

All algorithms presented were implemented in C++ and CUDA where applicable. All implementations were executed on a machine with an AMD R9 3900X 12-core CPU and an Nvidia GeForce RTX 3090 GPU. The authors intend to make source code publicly available, and apply for the Graphics Replicability Stamp (GRSI) \cite{grsi} upon publication.

Unless stated otherwise, the QUICCI descriptor resolution was set to 64x64 bits for all experiments. The support radius used was 100 units, which was found to be able to capture shapes in the local area. This trend is visible in the heatmap shown in Figure \ref{fig:occurrenceHeatmap}. All objects in the SHREC'16 dataset have been scanned at the same scale, and thus no scale alteration or correction was required.

\subsection{SHREC'16 Dataset Augmentation}
\label{subsec:augmentation}

The SHREC'16 Partial Retrieval track dataset \cite{Pratikakis_2016} has been chosen for the evaluation of the proposed retrieval pipeline. This allows direct comparison to results from other methods which were evaluated using this benchmark. While the dataset contains a variety of query objects, their quantity is limited, and is therefore augmented.

The used augmentation is similar to the one used in the SHREC'13 track for partial object retrieval \cite{sipiranSHREC13Track2013}; a first partial query set is created by generating meshes of all triangles in view from a random viewpoint. We used this to create one partial query mesh for each object in the SHREC'16 dataset, thereby creating 383 query objects. This query dataset is called $AUGMENTED_{Best}$. 

\begin{figure}
    \centering
    \includegraphics[width=8.0cm]{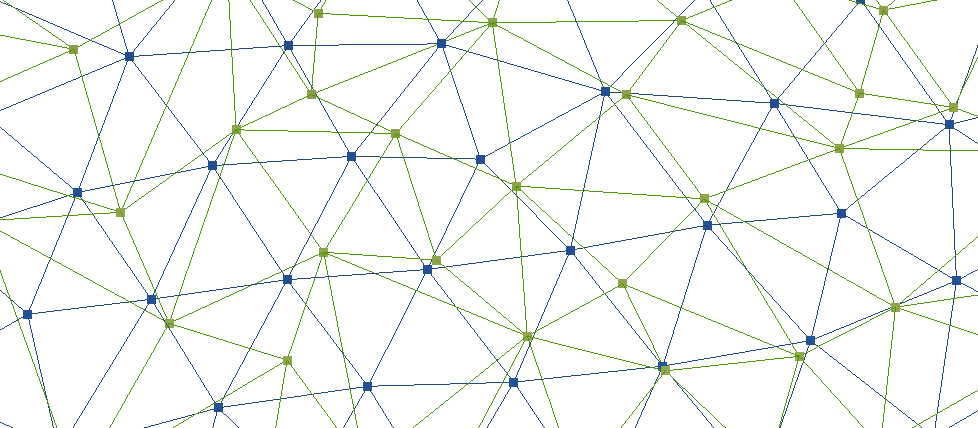}
    \caption{\small A part of two identical surfaces shown in wireframe form overlaid on top of one another, where one of the two has been remeshed. Vertices of both meshes have been highlighted.}
    \label{fig:surface_comparison}
\end{figure}

However, while the extracted query meshes give a good indication for \textit{best case} retrieval performance, a more realistic retrieval scenario could involve subsequent scans of the same object with different triangulations. We therefore also generated a second augmented dataset by remeshing all meshes in $AUGMENTED_{Best}$, creating the $AUGMENTED_{Rem}$ query dataset. In particular, we have used the remeshing algorithm proposed by Botsch et al. \cite{botsch2004a} as it works on non-watertight meshes, which has been made available as part of PMP library \cite{pmp-library}. An example of the effect of this remeshing step can be seen in Figure \ref{fig:surface_comparison}. 

The generated query datasets will be made available upon publication.

\subsection{Dissimilarity Tree}
\label{subsec:dissimilarity_tree_results}
The effectiveness of the dissimilarity tree index structure was evaluated by querying a tree constructed over all descriptors from all complete objects in the SHREC'16 dataset, which amounts to a total of 36.5M indexed descriptors. A set of 100,000 unique descriptors was randomly selected from the descriptors of all objects in the $AUGMENTED_{Best}$ set. 

\begin{figure}
    \centering
    \includegraphics[width=8.0cm]{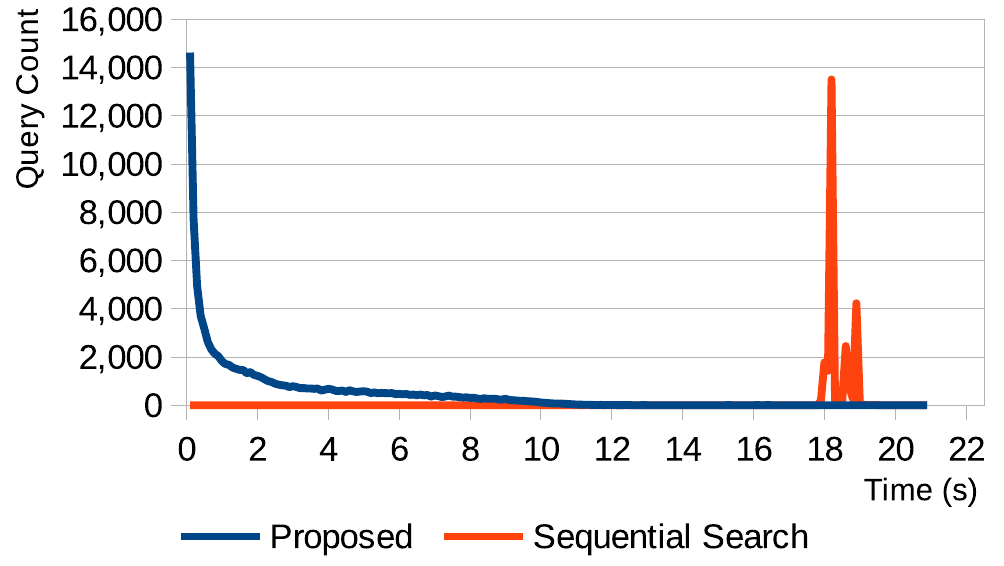}
    \caption{\small Histogram with bins of 0.1s showing the distribution of execution times of 64x64 bit queries using indexed and sequential searches. All sequential search occurrence counts have been scaled up by a factor of 10 for legibility.}
    \label{fig:index_query_execution_time_distribution}
\end{figure}

Each descriptor was subsequently used to query the tree. The resulting execution time of each query was counted in a histogram with bins of 0.1 seconds. As a reference, the first 2,500 queries were also used to measure the execution time of a sequential search, which resulted in another execution time distribution histogram. The results are shown in Figure \ref{fig:index_query_execution_time_distribution}.

The histograms show that significant speedups are achieved using the proposed dissimilarity tree structure over a sequential search. Out of the 100,000 queries, only 25 took longer than the average sequential search. 

The perf profiling tool showed that for a given query, on average 26.2\% of execution time is spent on visiting intermediate nodes, and 56.1\% is spent on visiting leaf nodes. The remainder is spent on open node queue management. Visiting leaf nodes is almost entirely (99.0\%) spent on the computation of weighted hamming distances. 

\subsection{Modified QUICCI evaluation}
\label{subsec:results_modified_equicci_filtering}

We evaluate here the effect of the modification to the QUICCI descriptor proposed in Section \ref{sec:quicci_modification}, which removes descriptor responses to object boundaries that typically exist in the partial query objects only. 

The $AUGMENTED_{Best}$ set is used to evaluate this modification. As each partial query object is extracted from a dataset object, there exists an exact correspondence between their vertices (ground truth). It is subsequently possible to determine the exact bits which are set in a query descriptor, but not in the correctly matching descriptor of the complete object, which are thus \emph{undesirable}.

\begin{figure}
    \centering
    \includegraphics[width=8.0cm]{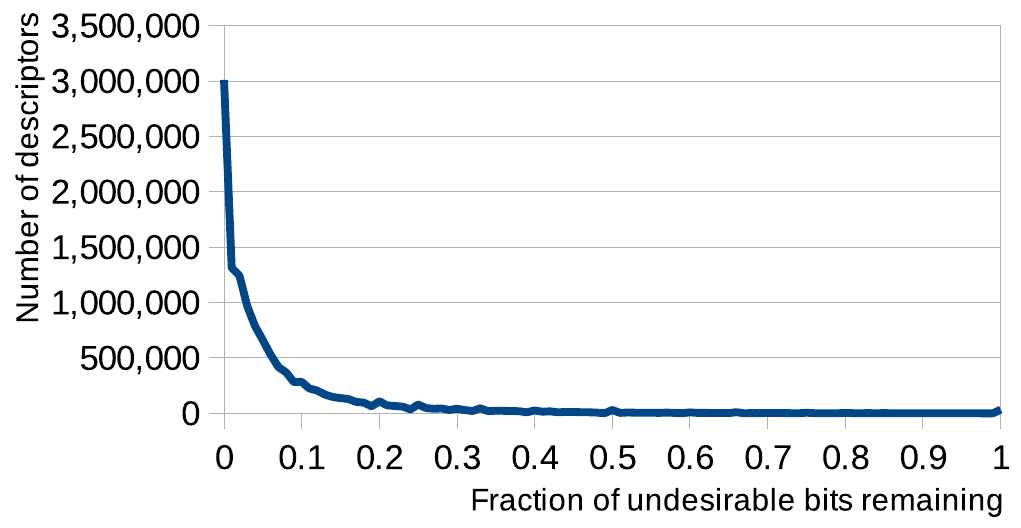}
    \caption{\small A histogram with bins of size 0.01, computed over 12.3M 64x64 bit partial query object descriptors, showing the fraction of undesirable (most boundary response) bits remaining in a modified QUICCI descriptor over the corresponding number of bits in the original QUICCI.}
    \label{fig:unwanted_bit_reduction}
\end{figure}

The chart in Figure \ref{fig:unwanted_bit_reduction} shows that in 78.0\% of the tested descriptors, the number of undesirable bits is reduced to under 10\%. In 24.5\% of all cases, the undesirable bits are removed entirely. The shown results have been computed over a set of 12.4M partial query descriptors, from which a relatively small number (105,448) have been excluded for not containing any query boundary responses (to avoid divisions by 0) or unreliable correspondence between vertices.

\begin{figure}
    \centering
    \includegraphics[width=8.0cm]{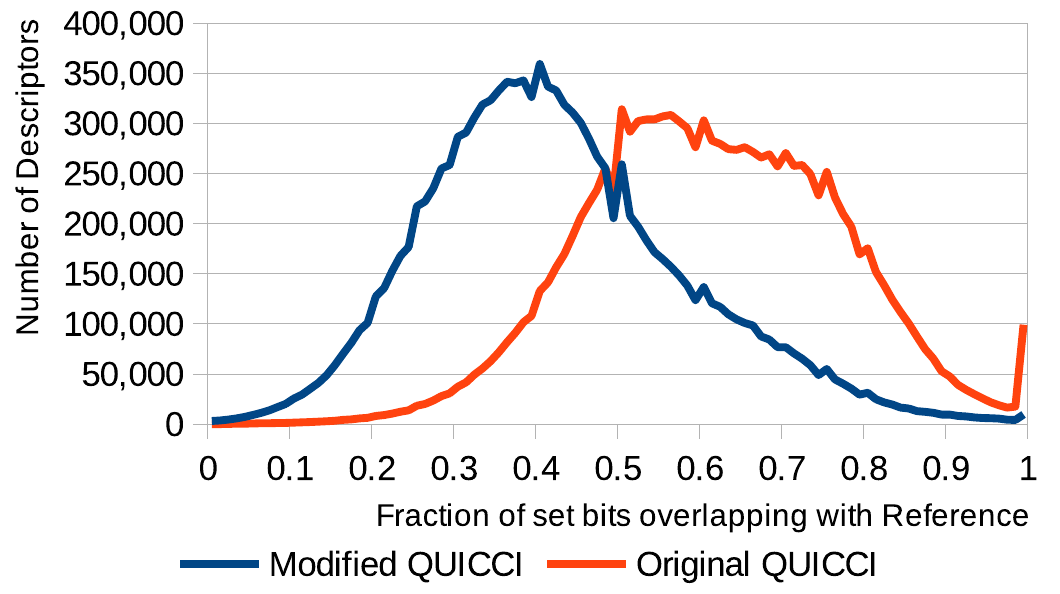}
    \caption{\small A histogram with bins of size 0.01, computed over 12.3M 64x64 bit partial query object descriptors, showing the distribution of fractional overlap of bits set to 1 in a partial query descriptor relative to the complete object from which the query was extracted.}
    \label{fig:overlap_with_reference_chart}
\end{figure}

However, the modification also removes some bits which are set in both the partial and complete descriptors, and are thus desirable. A histogram over the fraction of bits set to 1 in both descriptors relative to the total number of such bits in the complete descriptor is shown in Figure \ref{fig:overlap_with_reference_chart}. 

The Figure shows that the average of fractional overlap decreases from 61.5\% to 42.7\% when using the QUICCI modification, a loss of 30.6\%. However, while the fraction of desirable bits in query descriptors decreased, the average number of undesirable bits decreased even further, from an average of 84.2 bits per descriptor to 4.06 bits. Thus in the modified QUICCI, responses in the query descriptor can, to a high degree, also be expected to be present in the descriptor of the corresponding complete object.

\subsection{Modified QUICCI for Partial Object Retrieval}
\label{subsec:results_modified_quicci}

While Section \ref{subsec:results_modified_equicci_filtering} showed the proposed modification to the QUICCI descriptor to produce more reliable query descriptors, its effect on matching performance must be evaluated too.

A distance score for each of the query objects in $AUGMENTED_{Best}$ and $AUGMENTED_{Rem}$ was computed for each of the complete objects in the SHREC'16 dataset. The distance score of an object pair was computed by summing the distances of each descriptor in the query object to its nearest neighbour in the set of descriptors of the complete object. Next, all objects were ranked by their total distance to the query object. 
The results are outlined in Table \ref{tab:shrec2016_precision_chart}. As can be seen, the nearest neighbour matching performance improves significantly when using the modified QUICCI descriptors, for both the $AUGMENTED_{Best}$ and $AUGMENTED_{Rem}$ datasets.

\begin{table*}[]
\centering
\begin{tabular}{|l|l|l|}
\hline
\textbf{QUICCI} & \textbf{$AUGMENTED_{Best}$} & \textbf{$AUGMENTED_{Rem}$} \\ \hline
Original              & 0.72                     & 0.17                    \\ \hline
Modified     & 0.99                     & 0.49                    \\ \hline
\end{tabular}
    \caption{\small The fraction of correctly retrieved nearest neighbours when using combinations of the original and modified QUICCI, measured using the $AUGMENTED_{Best}$ and $AUGMENTED_{Rem}$ query object sets.}
    \label{tab:shrec2016_precision_chart}
\end{table*}

\begin{figure}
    \centering
    \includegraphics[width=8.0cm]{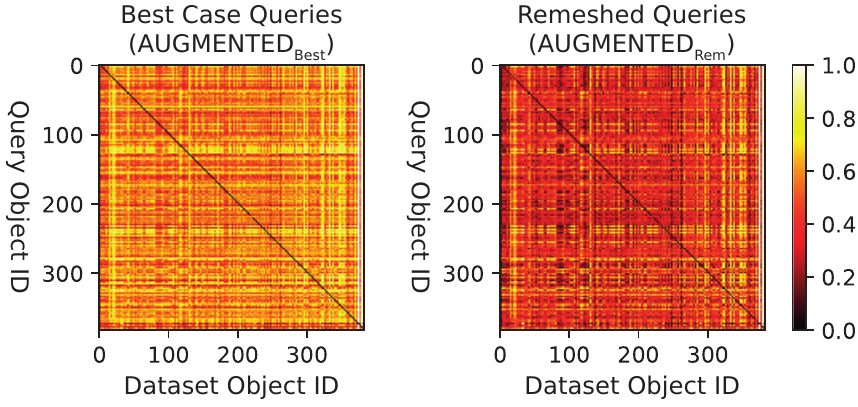}
    \caption{\small Confusion matrices showing summed Weighted Hamming distances from all objects in each of the augmented query object datasets to all objects in the SHREC'16 dataset. Since one partial query object is computed from each SHREC'16 dataset object, objects with matching IDs should correspond, and therefore have a low distance (leading diagonal). Distances of each row are normalised for legibility. Both query object datasets contain a visible desirable leading diagonal of low distances, although this is less pronounced for the remeshed query objects, which is also reflected in worse nearest neighbour matching performance.}
    \label{fig:confusion_matrices_chart}
\end{figure}

Confusion matrices were also computed across the query and complete objects, see Figure \ref{fig:confusion_matrices_chart}. Each row in these matrices represents the scores of a single query object to each complete object it was compared against. For each of these rows, the distance scores have been normalised to the range [0, 1].

The confusion matrix from the $AUGMENTED_{Best}$ set shows a clear distinction between partial query and complete objects. For the remeshed partial queries $AUGMENTED_{Rem}$, the nearest neighbour distance scores naturally increase. 

\subsection{Partial Retrieval Pipeline}
\label{subsec:results_pipeline}

Considering the proposed partial object retrieval pipeline, in this section we consider the nearest neighbour retrieval performance and the effect of the threshold parameter, as well as the execution times for querying objects.

\begin{figure}
    \centering
    \includegraphics[width=8.0cm]{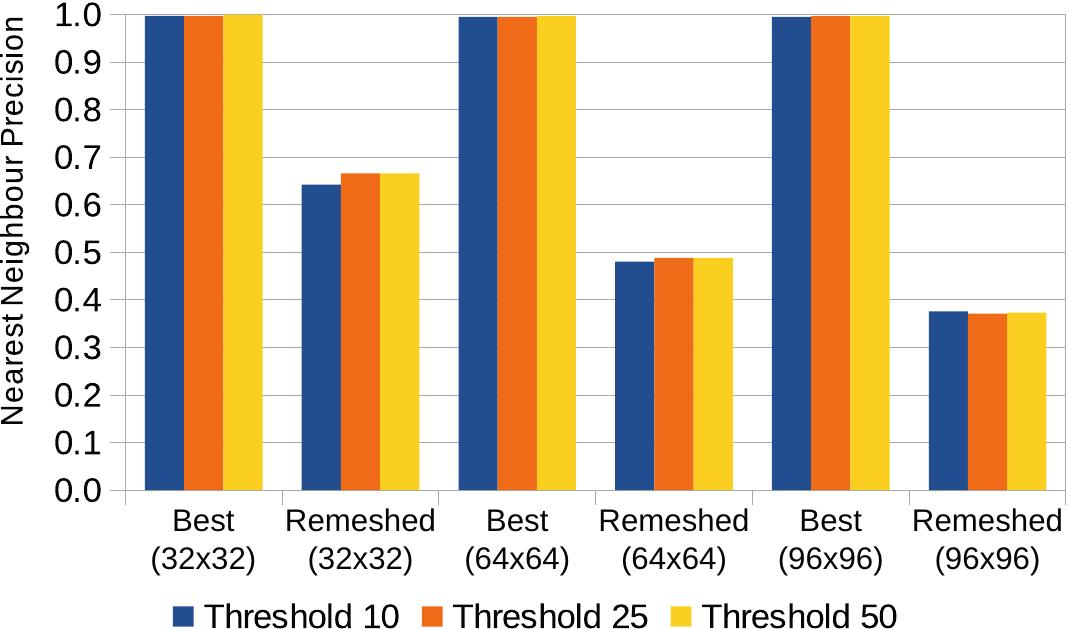}
    \caption{\small Fraction of correct nearest neighbour object matches for several voting thresholds, using the proposed partial object retrieval pipeline, tested using the $AUGMENTED_{Best}$ (Best) and $AUGMENTED_{Rem}$ (Remeshed) partial query object datasets.}
    \label{fig:objectSearch-results}
\end{figure}

Figure \ref{fig:objectSearch-results} shows the effect of the threshold parameter on nearest object neighbour retrieval performance for objects from $AUGMENTED_{Best}$ and $AUGMENTED_{Rem}$, for three different descriptor resolutions. As can be seen, the method is almost resilient to this parameter across resolutions and a low value can be used. 

While there is little variation in the results of query objects from $AUGMENTED_{Best}$, different descriptor resolution yields a significant variation in matching performance for the query objects of $AUGMENTED_{Rem}$.  

\begin{figure}
    \centering
    \includegraphics[width=8.0cm]{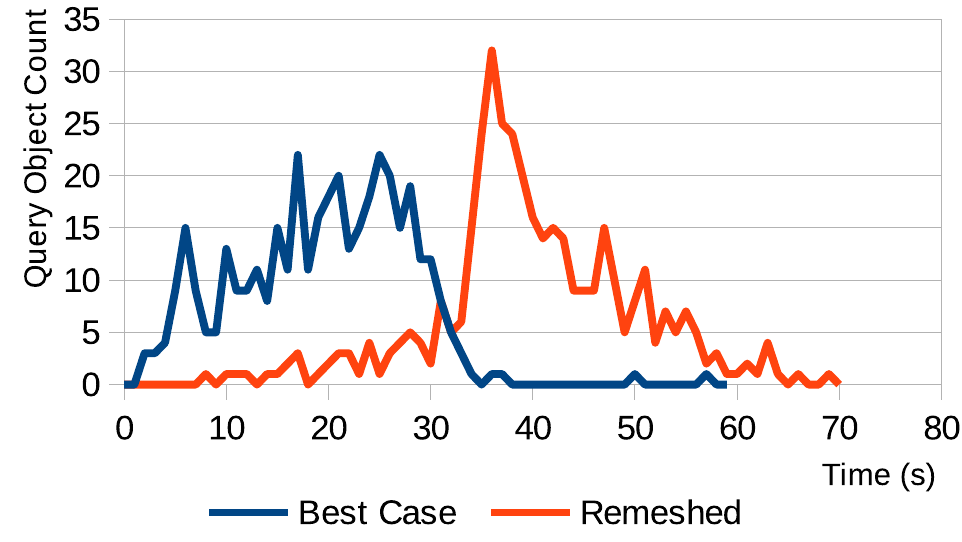}
    \caption{\small Execution times of the $AUGMENTED_{Best}$ and $AUGMENTED_{Rem}$ query meshes when using them in the proposed pipeline with a descriptor resolution of 64x64 and a vote threshold of 10.}
    \label{fig:objectSearch-execution-time}
\end{figure}

The execution times of the queries are shown in Figure \ref{fig:objectSearch-execution-time} using a vote count threshold of 10 and descriptor resolution of 64x64. It is also worth noting that queries based on different descriptors can be executed in parallel, although most of this acceleration was lost due to ensuring that results are reproducible. As can be seen in the Figure, there is a significant difference in the execution times of query objects from $AUGMENTED_{Best}$ and $AUGMENTED_{Rem}$. These results are discussed further in Section \ref{sec:Discussion}.

\subsection{SHREC'16 Partial Retrieval Performance}
\label{subsec:results_shrec2016}

The proposed partial retrieval pipeline is compared against the results presented in the SHREC'16 Partial Shape Query track \cite{Pratikakis_2016} as well as the results for the equivalent benchmark presented by Savelonas et al. \cite{savelonasFisherEncodingAdaptive2016}, which also includes results for the PANORAMA descriptor by Sfikas et al. \cite{sfikasPartialMatching3D2016} and Global Fisher features \cite{savelonasPartial3DObject2015}. As the source code of these works was not available, we have used the results from the referenced papers. Dimou et al. \cite{dimouFast3DScene2020} have also tested their work against this dataset, but no nearest neighbour retrieval performance was provided.

\begin{figure}
    \centering
    \includegraphics[width=8.0cm]{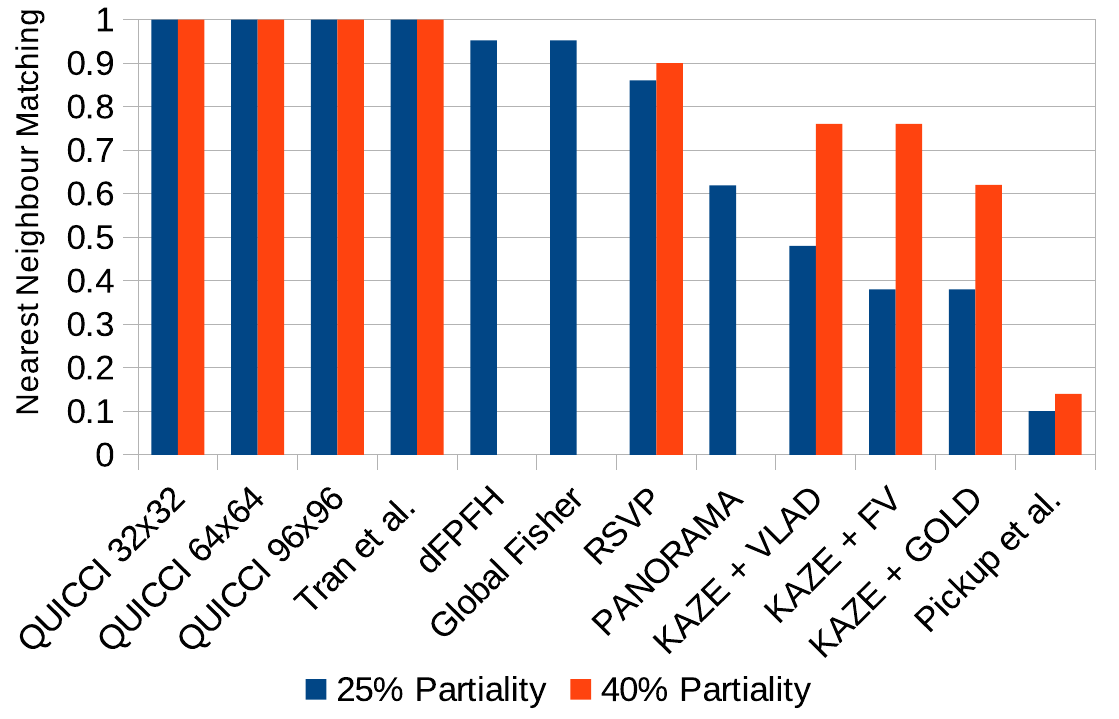}
    \caption{\small Comparison of nearest neighbour retrieval performance of the Virtual Hampson Museum collection. Query objects with 25\% and 40\% partiality were used. Results for the Tran et al., RSVP, KAZE+VLAD, KAZE+FV, KAZE+GOLD, and Pickup et al. methods are taken from \cite{Pratikakis_2016} and dFPFH, Global Fisher, and PANORAMA were taken from \cite{savelonasFisherEncodingAdaptive2016}. The latter does not show results for the 40\% partiality queries which are therefore missing.}
    \label{fig:shrec2016-results}
\end{figure}

The majority of the SHREC'16 benchmark focuses on the classification of objects into classes rather than specific object retrieval. Only the \textit{artificial queries}, which are culled versions of the database objects, have matching objects in the database. Fortunately, they also provide Nearest Neighbour data. Because the proposed retrieval pipeline is intended for exact part-in-whole matching, this comparison focuses on Nearest Neighbour. The results are shown in Figure \ref{fig:shrec2016-results}.

As shown in the Figure, the proposed method is able to correctly identify all partial queries in the benchmark, across multiple descriptor resolutions. While Tran et al. also accomplish this, their method uses the Iterative Closest Point (ICP) algorithm \cite{chenObjectModelingRegistration1991} \cite{beslMethodRegistration3D1992} 512 times per candidate match \cite{Pratikakis_2016}. While no execution times are listed, we estimate that our method is likely to run faster.

\section{Discussion}
\label{sec:Discussion}

\begin{figure}
    \centering
    \includegraphics[width=8.0cm]{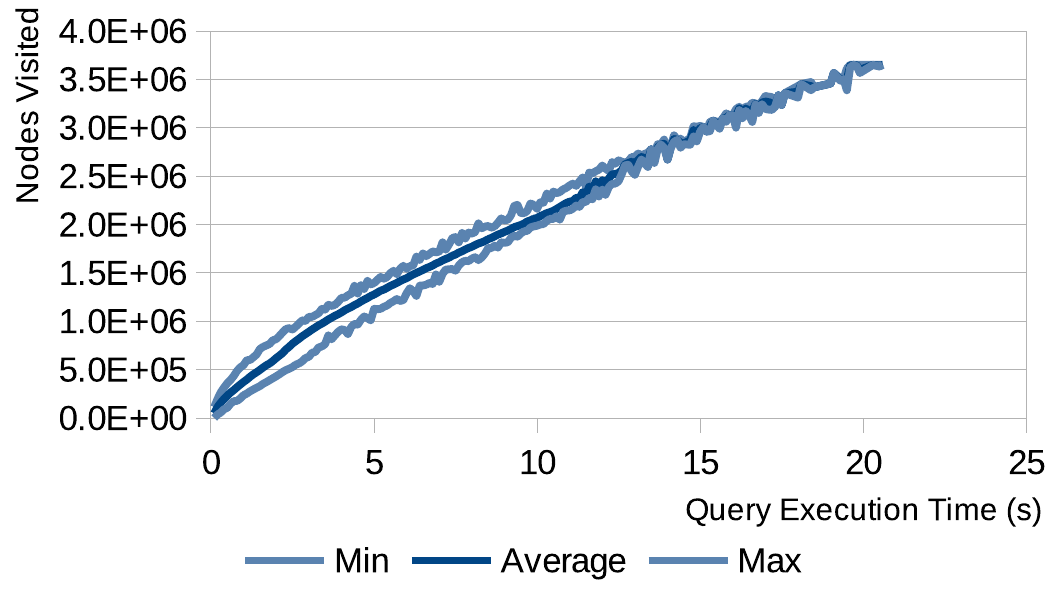}
    \caption{\small The relationship between the execution time of a query using the dissimilarity tree, and the number of tree nodes visited during that query's execution. For each time slice of 0.1 second, the minimum, maximum, and average number of nodes visited for all queries which executed within that time slice is shown.}
    \label{fig:node_count_visited}
\end{figure}

Figure \ref{fig:node_count_visited} shows that there appears to be a linear relationship between the query execution time and the number of tree nodes visited by the algorithm. As the querying algorithm iterates until it determines that no nodes with smaller distances than the ones found are present in the Dissimilarity Tree, one can conclude that the more dissimilar a query descriptor is from its nearest neighbour descriptors in the tree, the longer querying will take.

Figure \ref{fig:objectSearch-results} shows that there is a non-insignificant effect on partial object retrieval performance when using remeshed versions of the query objects. The severity of this effect varies across different descriptor resolutions. 

The cause of this loss in matching performance is that the positions of vertices on the object surfaces are slightly shifted causing changes in intersection counts to occur elsewhere on the descriptor, which can be seen in Figure \ref{fig:surface_comparison}. 

Because the distance between each vertex and its closest neighbour in the remeshed mesh is small, and based on anecdotal evidence, the resulting effect is that corresponding QUICCI descriptors on the unmodified and remeshed object contain portions of bits which have either been shifted left or right by 1 bit. As the Weighted Hamming distance function only considers bits in exactly the same position, this incurs a distance penalty on what would otherwise have been a good match. 

The probability of this distance penalty occurring is diminished when the descriptor resolution is lowered. As the distances between QUICCI intersection circles is increased to cover the same support radius, the probability of bit shifts decreases, thereby resulting in the improved matching performance observed in Figure \ref{fig:objectSearch-results}.

The distance penalty also has the downside of increasing query execution times. As indicated in Figure \ref{fig:node_count_visited}, we observed a relationship between the similarity of a query descriptor and the nearest neighbour in the set of complete object descriptors, and the execution time of that query. When the distance to the nearest neighbour increases, so does the execution time of the dissimilarity tree search algorithm. This increase is visible in Figure \ref{fig:objectSearch-execution-time}.

Given the extremely promising nature of the retrieval results of the partial query objects from the $AUGMENTED_{Best}$ dataset, we conjecture that if in future work a distance function is found which can remedy the aforementioned distance penalty issue, it should be possible to both significantly increase the matching performance of remeshed queries while simultaneously reduce query times.

\section{Conclusion}

A small modification to the QUICCI descriptor was shown to be advantageous for partial retrieval tasks. An indexing scheme for binary descriptors called \textit{Dissimilarity Tree} was also proposed, and was shown to greatly reduce nearest neighbour retrieval time. Finally, an accurate and efficient search algorithm for partial 3D object retrieval using the aforementioned Dissimilarity indexing structure was proposed.

\section{Acknowledgements}

The authors would like to thank the HPC-Lab leader and PI behind the "Tensor-GPU" project, Prof. Anne C. Elster, for access to the Nvidia DGX-2 system used in the experiments performed as part of this paper. Additionally, the authors would like to thank the IDUN cluster \cite{sjalander+:2019epic} at NTNU for the provision of additional computing resources. Multiple images in this work were captured using Meshlab \cite{meshlab}.

\printbibliography

\end{document}